\documentclass[10pt, openany]{book}

\usepackage[]{graphicx}
\usepackage{booktabs}
\usepackage[]{color}
\usepackage{alltt}
\usepackage[T1]{fontenc}
\usepackage{adjustbox}
\usepackage[utf8]{inputenc}
\usepackage{threeparttable}
\usepackage{graphicx}
\usepackage{amsmath}
\usepackage{lipsum}
\usepackage{color}
\usepackage{rotating}
\usepackage{multirow}
\usepackage{hhline}
\usepackage{wrapfig}
\usepackage{caption}
\usepackage{hyperref}
\usepackage{cite}
\usepackage[table,xcdraw]{xcolor}

\usepackage{tabularx}

\usepackage[top=100pt,bottom=100pt,left=68pt,right=66pt]{geometry}

\usepackage{lipsum}

\usepackage[swedish, english]{babel}

\usepackage{fancyhdr}
\usepackage{titlesec}
\titleformat{\chapter}
   {\normalfont\LARGE\bfseries}{\thechapter.}{1em}{}
\titlespacing{\chapter}{0pt}{50pt}{2\baselineskip}

\usepackage{float}
\floatstyle{plaintop}
\restylefloat{table}

\usepackage[tableposition=top]{caption}

\usepackage{hyperref}
\hypersetup{hidelinks,linkcolor = black} % Changes the link color to black and hides the hideous red border that usually is created

\graphicspath{ {images/} }

\frontmatter

\begin{document}

%%% Selects the language to be used for the first couple of pages
%\selectlanguage{swedish}

%%%%% Adds the title page
\begin{titlepage}
	\clearpage\thispagestyle{empty}
	\centering
	\vspace{2cm}

	% Titles
	{\large Computer Networks - COEN 329 \par}
	\vspace{4cm}
	{\Huge \textbf{Evolution of Artificial Intelligent Plane}} \\
	\vspace{1cm}
	{\large \textbf{Artificial Intelligence Plane} \par}
	\vspace{4cm}
	{\normalsize Puneet Kumar \\ % \\ specifies a new line
	             SCU-ID: 00001424550 \par}
	\vspace{2cm}

    \includegraphics[scale=0.60]{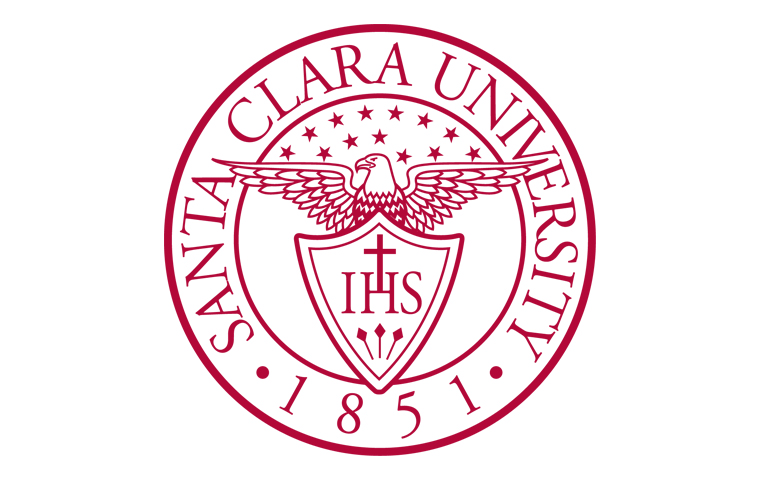}
    
    \vspace{2cm}
    
	% Information about the University
	{\normalsize Santa Clara University \\ 
		Computer Networks: COEN-329 \\
		Department of Computer Science \par}
		
	% Set the date
	{\normalsize California, USA \par}
	\vspace{2cm}
	
	\pagebreak

\end{titlepage}

\chapter{Audience}
\label{chapter_1}
\large This document is intended to explain the new Artificial Intelligent Plane in addition to control, management and data plane, overlooking the three planes. 
Primary objective of this new plane is to bring centralized management, ease the data path and minimize traffic for control plane. With the tremendous growth of internet, scaling of networks requires new intelligent ways to make networks smart. This document targets audience of the class who are taking taken COEN - 329 (Network Technology), as fundamentals to understand this document was explained in the class. In lectures we covered several networking techniques in L2/ L3 and above networking layers and explained control, data and management plane.

With the growth of the internet, it is becoming hard to manage, configure and monitor networks. 
Recent trends to control and operate them is artificial intelligence based automation to minimize human intervention.
%To control and operate %huge networks, them recent trends in networking is artificial intelligence based automation to minimize human intervention. 
Albeit this concept has been introduced since a decade with several different names, but the underlying goal remains the same, which is to make network intelligent enough to assemble, reassemble if configuration changes, and detect a problem on its own and fix it. As a result, in addition to \textit{Data Plane, Control Plane and Management Plane}, a new plane called \textit{Artificial Intelligence (AI) Plane} is introduced.

Our main objective is to analyze all major AI plane techniques, frameworks and algorithms proposed in various types of networks. We propose a comprehensive and network independent framework to cover all aspects of AI plane, in particular we provide a systematically means of comparison. In conjunction to make AI plane understand simpler, this framework highlights relevant challenges and design considerations for future research. To the best of our knowledge this is the first survey report which represents a complete comparison of AI planes with their investigation issues in several types of networks.

% Adds a table of contents
\tableofcontents{}
%\clearpage
\listoffigures
%\clearpage
\listoftables

%%%%%%%%%%%%%%%%%%%%%%%%%%%%%%%%%%%%%%%%%%%%%%%%%%%%%%%%%%%%%%%%%%%%%%%%%%%%%%%%%%%%%%%%%%%%
%%%%%%%%%%%% The rows above should not be changed except for the title page information
%%%%%%%%%%%%%%%%%%%%%%%%%%%%%%%%%%%%%%%%%%%%%%%%%%%%%%%%%%%%%%%%%%%%%%%%%%%%%%%%%%%%%%%%%%%%
%%%%% Text body starts here!
\mainmatter

\chapter{\large Introduction}
\label{intro}
\large Networks are evolving to meet user demands. 
Main qualities which make conventional internet successful are heterogeneity and generality combining with user transparency and rich functionality for end-to-end systems. 
In today's world networks display characteristics of unstable convoluted systems. 
Till date most networks are murky to its applications and providing only best effort delivery of packets with little or zero information about the reliability and performance characteristics of different paths. 
Granting, this design works well for simple server-client model, many emerging technologies such as: NFV (Network Function Virtualization \cite{li2015software}, IoT (Internet of Things) \cite{atzori2010internet}, Software Defined Networking \cite{mckeown2009software}, CDN (Content Delivery Networks) \cite{bertrand2012use} and LTE (Long-Term Evolution) \cite{sesia2011lte} and 5G Cellular Networks \cite{ggg} heavily depend on affluent information about the state of the network. 
For example, author in \cite{NFVStateOfNetwork} described, if VNFs (Virtual Network Functions) \cite{VNFDefinition} are not aware of the traffic on virtio interfaces  assisting hypervisor, then this might result in a bottleneck in NFV infrastructure. 
In other words, VNFs should know the state of the network (in terms of traffic) to accelerate applications hosted across VNFs in NFV infrastrucutre. 
Authors in \cite{IoTStateOfNetwork} explained the need of the data storage as the number of connected IoT devices are increasing on unprecedented level \cite{evans2011internet}. 
In order to optimize the data storage, it is imperative for IoT nodes to know about the other nodes and their transportation method of moving data among networks. 
Literature in \cite{SDNArtificialIntelligence} specifically pinpointed SDN problem of not knowing state of the network when dealing with migrating VMs. 

\begin{figure}
    \centering
    \includegraphics[width=0.8\linewidth]{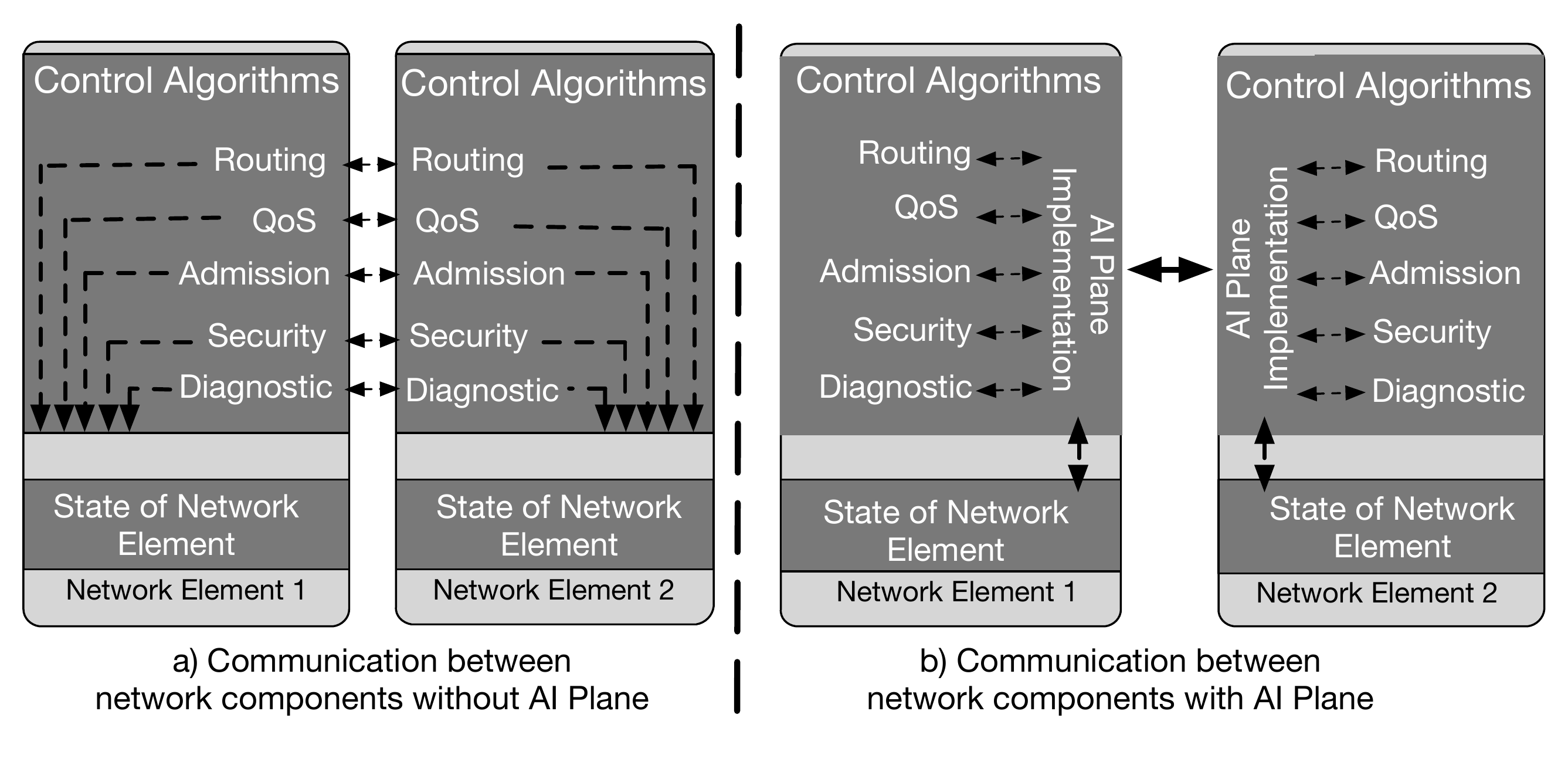}
    \captionsetup{font=scriptsize}
    \caption{\textbf{Network with AI and non-AI plane}}
    \label{fig:NewNewAiPlane}
\end{figure}

Conventional internet is known for its mercurial nature due to several inter-winded connected devices. These connected devices designed, deployed, and optimized individually to work together based on network protocols. 
Notwithstanding these devices demonstrate feasibility of enlarging new innovative services but they fail to gather and retain information methodically useful for the network as whole. 
At present conventional algorithms in networks are incompetent to sophistication of behavior due to software and protocols coordinating network element's \footnote{Any device which has network function such as routing, switching etc is called network element} control and management planes, specially how the decision logic and the distributed-systems issues are inescapable entwined. 

\begin{figure}
    \centering
    \includegraphics[width=0.60\linewidth]{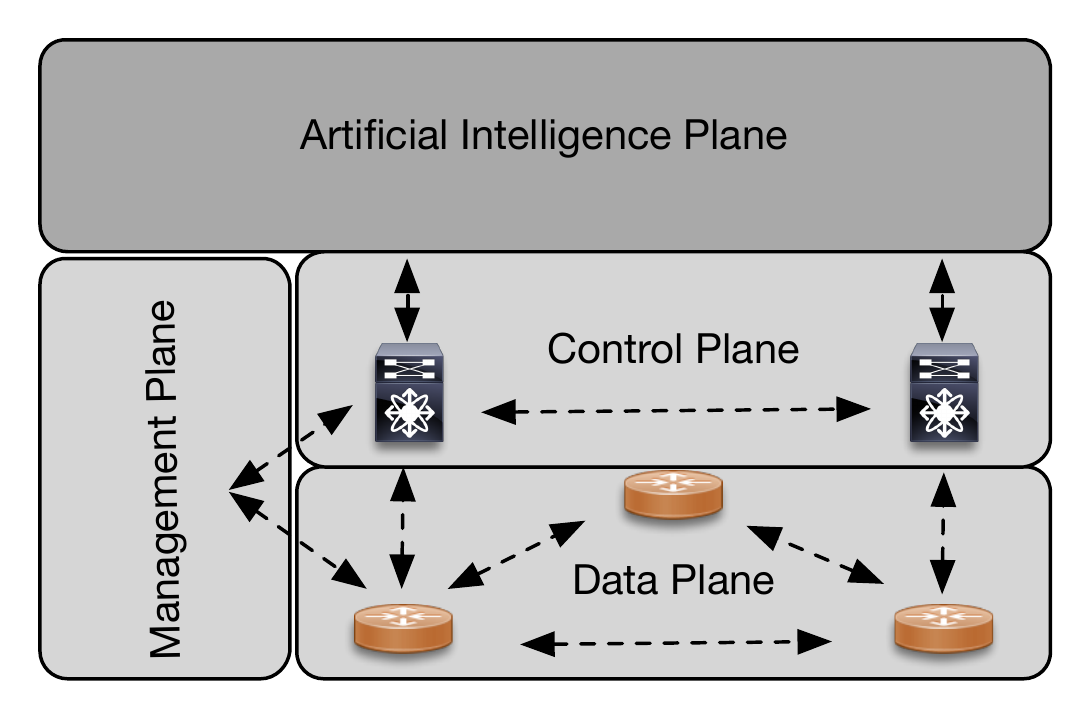}
    \captionsetup{font=scriptsize}
    \caption{\textbf{Interaction of AI plane with other planes}}
    \label{fig:AI_plane}
\end{figure}

In order to solve these issues, a new plane was visioned by \textit{Clark at el.} called \textit{Knowledge Plane} (KP) \cite{clark2003knowledge}. According to Clark at el., KP is based on intelligent learning and may shift the paradigm on the methods we presently operate, troubleshoot and optimize computer data networks. 
Although AI plane has a lot of similarities with KP plane, but over a period of time it has evolved a lot.
As Fig. \ref{fig:AI_plane} depicts, AI plane is an addition to the three traditional planes as we know today: \textit{\textbf{Management Plane, Control Plane and Data Plane}}. 
It decouples decision logic from distributed protocols and enables simpler protocols to operate in these three traditional planes.
Management plane deals mainly with supervision and configuration of the network, Data Plane is liable for forwarding, processing and in some cases storing of data packets. Control Plane assists Data Plane for matching and processing rules. 
AI plane duties in broader perspective are to obtain a rich view of network, learn the behavior of network, and if possible then to operate on them.

Fig \ref{fig:NewNewAiPlane} shows two communication model with and without AI plane. 
Figure a) shows the communication model between two network elements without AI plane. Each control mechanism has to glean information from hardware for its own use. Each control mechanism shares this information with the equivalent entity in other network. b) Shows the communication model with AI Plane. Here, AI plane shares and gathers information from each control mechanism to provide germane and rich information to other network element's AI plane.
Each control algorithm \footnote{Trafﬁc control, QoS architectures, differentiation of services etc.} has to attain information from the hardware for its own use. This control information is exchanged between corresponding algorithm of other network. Hence, redundancy between different control algorithms is inevitable and makes it hard to design global information management. For example, every control algorithm assuredly needs to know whether the direct neighbors are alive or not. Such information is not shared by different algorithms in such architecture. Another example is load on interfaces, every control algorithm might be interested in such information. Without AI plane, each algorithm has to query its own control information from hardware, represent and use it by itself. Although this might work in usual network control plane but this is definitely not scalable. On the other hand, AI plane takes responsibility for gathering and sharing data. It eliminates the need for every control algorithm to query hardware and can pre-compute and correlate data in a rich format. Such augmented information is called \textit{Knowledge}. 

Our research focuses on the following contributions: First, a comprehensive framework is proposed for design AI plane. 
The emphasis of this framework is to aid in comparison and describe the focal point of existing AI plane algorithms. 
Although, previous research focuses on developing and designing AI planes for various networks. 
Our focus would be to understand those AI planes and their applications in detail. 
Secondly, all types of AI planes will be covered irrespective of their networks type, including their limitations and merits. 
Finally, we will discuss future research and untapped areas in AI plane. 

\chapter{\large AI Plane Framework}
\large  AI plane paradigm is about optimizing networks by providing automation, recommendations, estimations and predictions. Theoretically, this paradigm acquired few concepts from other ideas such as: Derivative-free-optimization \cite{rios2013derivative}, Self-Organizing Systems \cite{SelfOrganizingMobileCommerce}, Ego-Centric Context Aware Ad Hoc Networks \cite{julien2002egocentric}, feedback control system based on neural networks \cite{narendra1990identification} and autonomic network management architecture \cite{derbel2009anema}. 
Due to the diversity in different type of networks with various goals, settings and implementations, it is arduous to make an all-inclusive comparison among AI planes. 
This emanates a need for comprehensive framework to make comparison smooth between AI planes. In this section, a framework is proposed to ease comparison in AI planes and the building block components of it.

\begin{figure}
    \centering
    \includegraphics[width=0.60\linewidth]{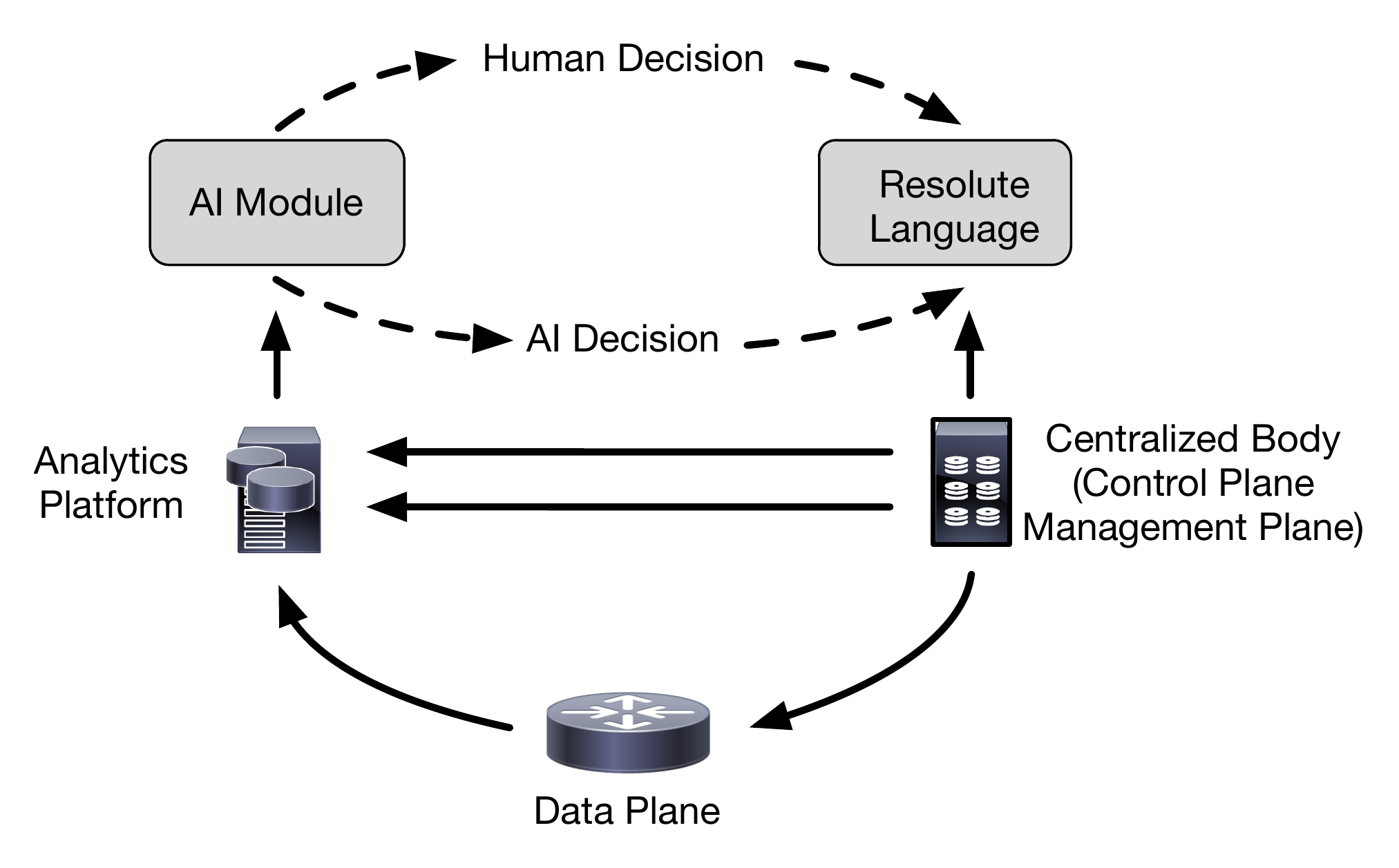}
    \captionsetup{font=scriptsize}
    \caption{\textbf{Framework to compare AI Planes}}
    \label{fig:AIplaneFramwork}
\end{figure}

Fig. \ref{fig:AIplaneFramwork} shows the framework, which is based on the shared features of the AI plane. Building blocks of the framework are described below.

\section{Analytics Module} 
\begin{figure}
    \centering
    \includegraphics[width=0.60\linewidth]{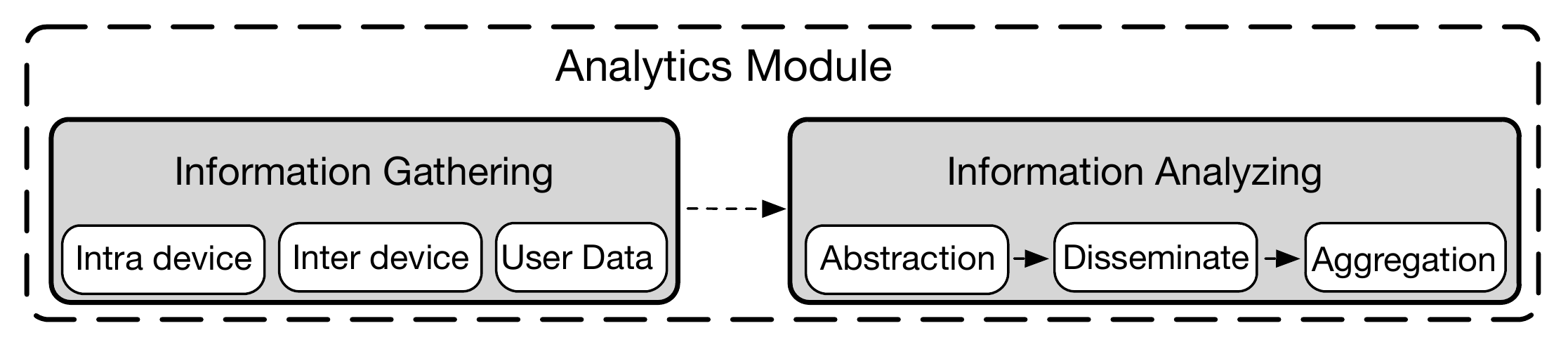}
    \captionsetup{font=scriptsize}
    \caption{\textbf{Gathering and analyzing of information}}
    \label{fig:AnalyticsModule}
\end{figure}

As depicted in Fig. \ref{fig:AnalyticsModule} this module has dual tasks. First, it gathers information about state of the network either from network elements or from other systems. Second, it analyzes the gathered data. 
Information gathered by AI plane is widely distributed and available to be processed sophisticatedly and adaptively by \textit{Analytics Module} (AM). 
AM has two main duties: a) Information gathering b) Analyzing Information. Gathering information heavily depends on the final objective of \textit{Artificial Intelligence} (AI) module. For instance, if AI needs to send decision cross layers in network protocol stack then AM would collect information from other layers as well. If AI needs to send decision based on the system level network elements such as dynamic routing, switching etc, then AM needs to collect data from other devices to have a complete picture of network. 
There are several types of data can be collected but for the scope of this report, we have categorized into three types:

\textit{1) Intra device:} This kind of information gathering for optimization happens in a single device. For example, protocol parameters for optimizing routing or location for security or energy conservation.

\textit{2) Inter-device:} Information exchanged among devices are called \textit{inter device} information gathering. For instance, traffic patterns or queue lengths in routers to optimize the maximum capacity of flows a system can serve.

\textit{3) User data:} This depicts user preferences. For instance state the Quality of Experience to enhance QoS in cellular access networks.

On the other hand, while analyzing information, AM studies the granularity of data and define practical methods for representing and retrieving such data at  device or system level. Main components of analyzing data are:

\textit{1) Abstraction:} Since information gathered can be from different elements within or outside network, in order to give meaning to the data with respective to the entity it belongs to, AM abstract the data before it starts analyzing it. Please note that it is important to clean and process raw data prior to abstraction by processes such as discretization \cite{Discretization}, normalization \cite{normalization} and missing value completion. For example, TCP \cite{allman2009tcp} in network stack coming from different sources can be abstracted to be analyzed.

\textit{2) Disseminate:} After abstraction, information is widely dispersed among respective entities to be analyzed. For instance, to continue the example of TCP connections given above, network parameter's (RTT (Round Trip Time), protocol, Qos etc) information will be effectively dispersed to each entities to be analyzed. RTT and inter-arrival time \cite{feldmann2000characteristics} will be analyzed to help the best size of the TCP congestion window \cite{winstein2013tcp}. In artificial intelligence paradigm, discovering and disseminate proper features is the key to fully understand the potential of the data.

\textit{3) Aggregation:} This part of the AM module is responsible to aggregate the analyzed result to be sent over to AI module to make an intelligent decision. 
%In example given above, policies based on traffic pattern or forwarding decisions will be sent to AI module to execute.

\section{Artificial Intelligence Module} 
Since decision making is the most challenging part of any machine learning process which often requires efficient and adequate data analysis, makes AI module (depicted in \ref{fig:AIModule}) the most important component in the framework.
Present and prior data provided by AM module are fed to the learning algorithms to generate the output. There are three famous approaches for deep learning \cite{lecun2015deep} depending on the nature of learning objects: a) Administered Learning b) Un-administered Learning c) Prime Learning. Their descriptions are given below:

\begin{figure}
    \centering
    \includegraphics[width=0.75\linewidth]{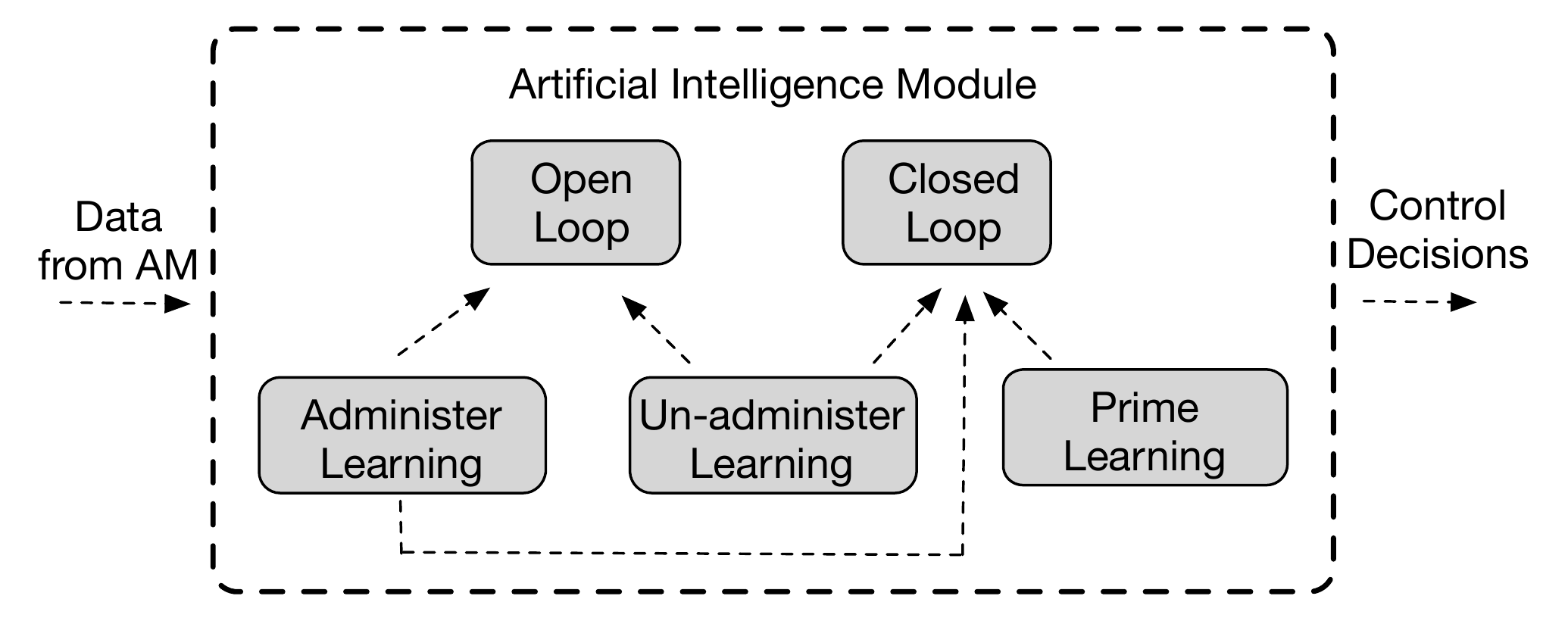}
    \captionsetup{font=scriptsize}
    \caption{\textbf{Figure depicts AI plane. }}
    \label{fig:AIModule}
\end{figure}

\textit{1) Administered Learning:} Example inputs with their desired outputs are fed to administered learning function, with an intent to come up with a general rule that maps inputs to outputs. For instance, administered learning has been widely applied to wireless cellular networks in channel estimation \cite{huang2018deep}.

\textit{2) Un-administered Learning:} Unlike administered learning, Un-administered learning function should be able to find its own embedded structure or pattern in its input. Usually un-administered algorithms tends to find hidden patterns and find a suitable representation in fed data. For example,  Singular Value Decomposition (SDA) \cite{de1994singular} and Principal Component Analysis (PDA) \cite{jolliffe2011principal} are used to manipulate the receiving matrix of massive MIMO \cite{larsson2014massive} in order to reduce the computational complexity.

\textit{3) Prime Learning:} Prime learning function obtains its objective by interacting with dynamic environment. This type of learning is inspired by \textit{control theory} \cite{simrock2008control} and \textit{behaviorist psychology} \cite{BehaviorPsychology}, however entity who runs it, doesn't have explicit knowledge of whether it has come near to the final objective or not. Entity should take actions in an environment to maximize the aggregated in Markov Decision Process \cite{MarkovProcess}. As an example, user sets a target policy, say delay in a set of flows, then entity acts on the centralized body (SDN Controller may be) by altering the configuration, every action receives reward, which increases as present policy gets closed to the target policy. Prime Learning specifically has provided some extraordinary results, notable mentions are \cite{mnih2015human, silver2016mastering}.

AI plane smooths the transfiguration between measured data collected by the AM module and control specific actions. Usually network administration has to examine all the network parameters, figure out the metrics which interest the final goal and make a decision to achieve that goal. This process will be handed to AI plane, which will be able to make or recommend control decisions with the help of AI module. AI module expresses those control decision with the help of resolute language. This heavily assists the transition between AI Module decisions and low level decisions made by data, management and control plane elements. Depending on the learning approaches defined above, there are two different sets of application for the AI plane:

\textit{1) Closed Loop:} Network model obtained by this set of applications can be used in two cases. First is automation, AI module can make decisions automatically based on behalf of the network operator. 
Second for optimization of the existing network configuration, subjected upon the condition if learned network model can be analyzed by common optimization techniques to find optimal configuration (quasi). 
Both of those cases can also be done by prime, administered and un-administered learning through the intent interface provided by control or management plane of the controlled body as shown in Fig. \ref{fig:AIplaneFramwork}.

\textit{2) Open Loop:} In this set of applications, network administration conducts the decision making process, nevertheless AI module can ease the task. In case of administered learning, network model learned can be used for validation. A network administrator can query the network model to validate the provisional changes before committing them to system. In un-administered learning, recommendations are provided to network administrator based on the correlations found in the explored data.

\section{Resolute Language} A common way is required to express the intent from both network administrator and AI Module on their behalf. This module serves a common interface to both human or AI module interactions and defines accurately how the intent should be translated into specific control decisions. It should be noted that this is different than the language used by low level network elements sitting in Control, Management and Data plane. Some references of resolute language has already been discussed in \cite{foster2013languages, kim2013improving}.

\chapter{Artificial Intelligence Planes}
Over the past decade AI Planes have been presented in different flavors, usually in meta-control plane (i.e control algorithms being controlled by other algorithms). 
This idea sounds very captivating because in broad perspective, it would cover autonomic networking field: managing high-level knowledge and making self-managed networks. 
But the success of new machine learning approaches has evolved AI plane as a critical field in networking, earning an entire field by itself. 
There are two types of AI planes, first is the type which focuses on system level optimization, while second's final objective is to optimize entire network. 
Irrespective of AI plane's network types, our comparison is based on the factor whether AI plane is used to optimize the whole network or a single system.

\begin{figure}[t]
    \centering
    \includegraphics[width=0.65\linewidth]{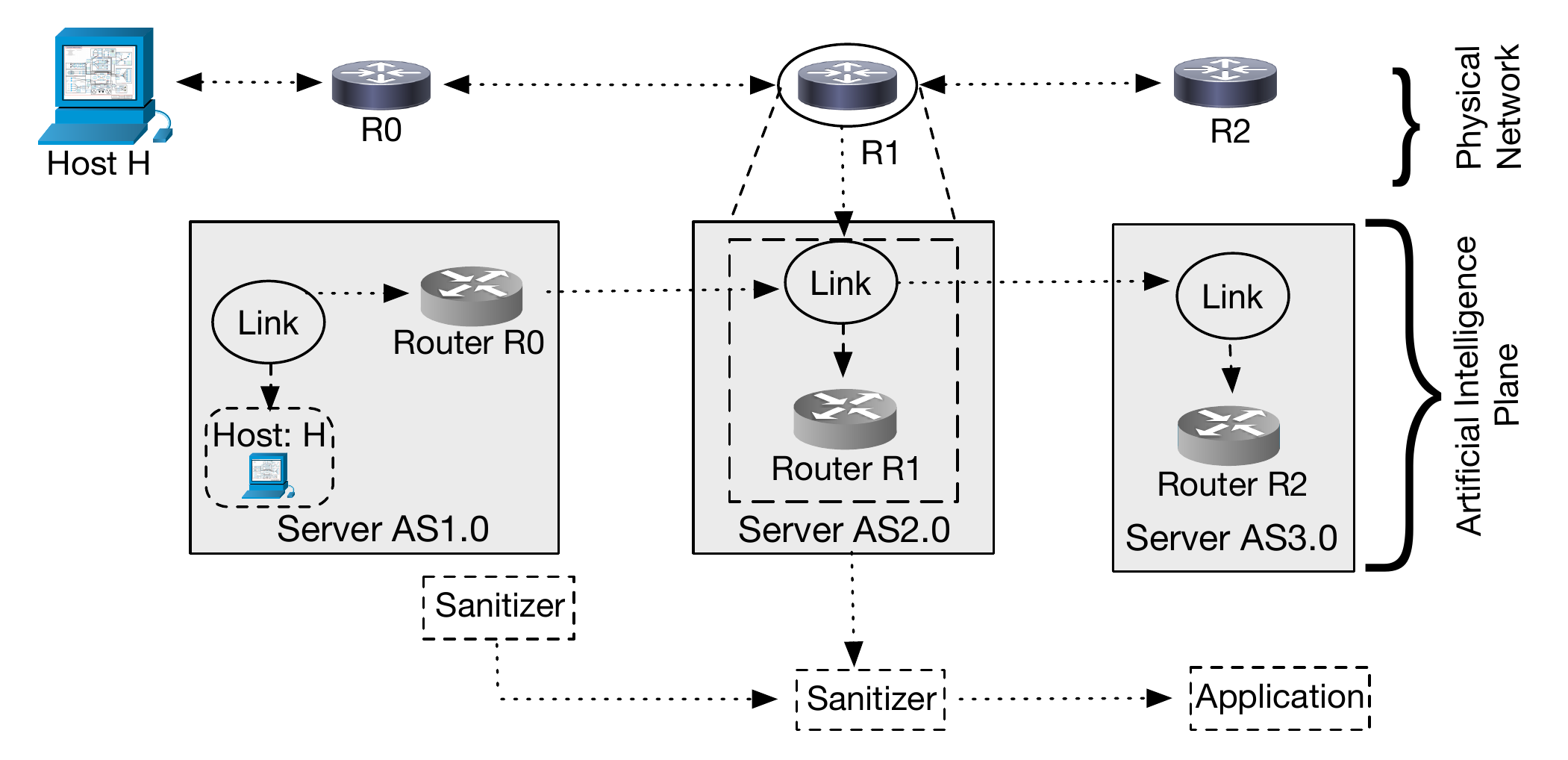}
    \captionsetup{font=scriptsize}
    \caption{\textbf{Dissemination of information}}
    \label{fig:NetQuery}
\end{figure}

\section{System Level AI Plane}
These type of AI planes primarily focus on system level optimization. Although their final goal might be optimizing the entire network but primarily they are focused on optimizing one system only.

\begin{figure}[t]
    \centering
    \includegraphics[width=0.6\linewidth]{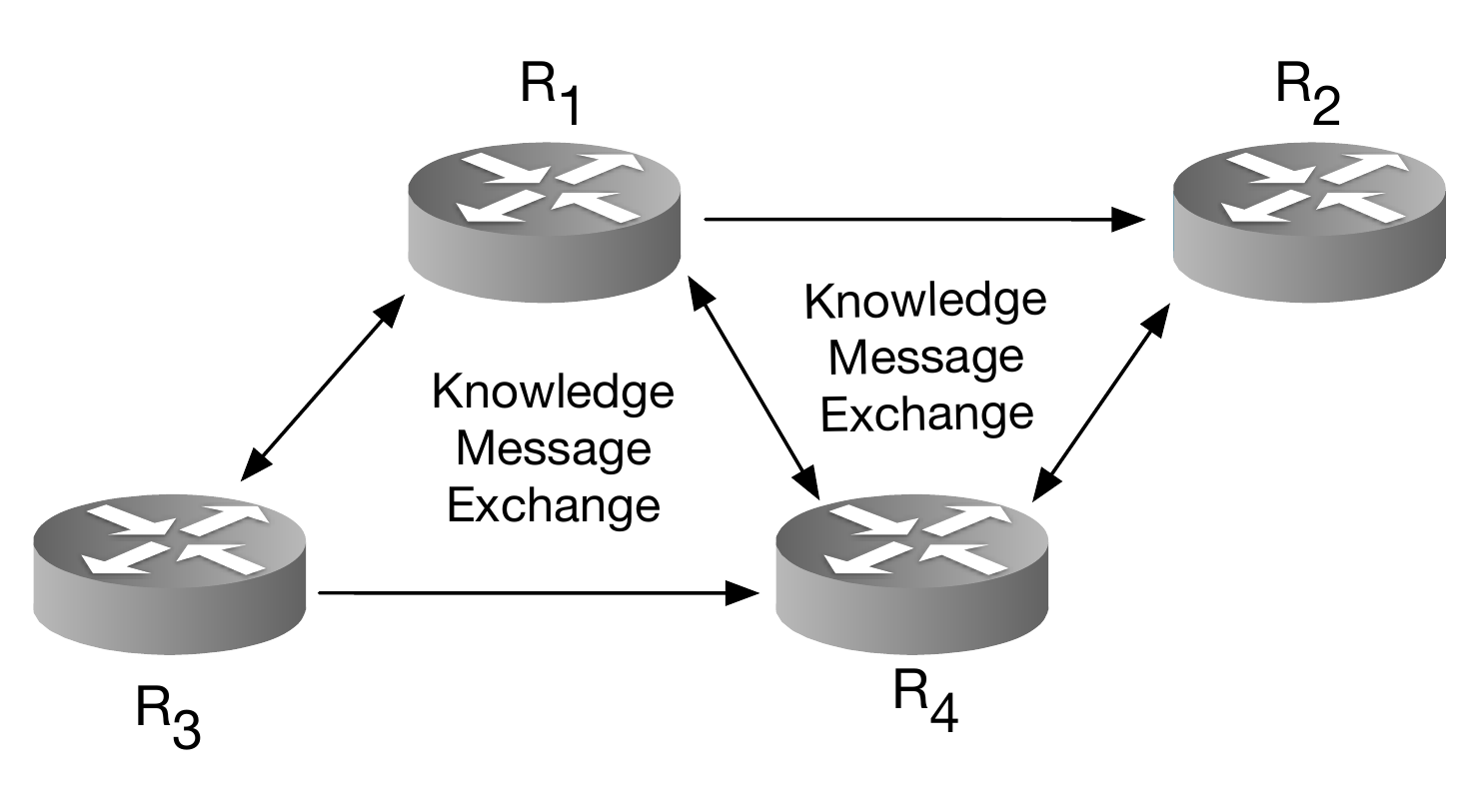}
    \captionsetup{font=scriptsize}
    \caption{\textbf{Exchanging knowledge messages}}
    \label{fig:SituatednessBasedKP}
\end{figure}

\subsection{Situatedness-Based Knowledge Plane \cite{bullot2008situatedness}} 
AI plane mentioned in \textit{Situatedness-Based Knowledge Plane} is a derivative of autonomic computing \cite{kephart2003vision}. 
The autonomic computing initiative doesn't concentrate on a collection of homogeneous network elements, rather on intelligent and dedicated information systems made of heterogeneous network elements, where each network element can  have its own knowledge management including information gathering and analyzing.
With the increase in number of users, more network elements are being introduced and so the services associated with it. 
These services engendered new intricacies of layers and govern these layers is moving beyond network administrator's control. 
To address this issue Situatedness-Based Knowledge plane uses \textit{collaborative and autonomous agents} aka Multi-Agent Systems (MAS) \cite{gattinetwork, merghem2003using}. These agents are ingrained in network element itself, which serves and deliver local and situatedness knowledge comprising the Situatedness-Based Knowledge Plane.

As mentioned earlier, Situatedness-Based Knowledge plane is based on MAS, so first question arises: what is MAS? How do network or network element fit into this category? Authors in \cite{bieszczad1998mobile} defined MAS as a structure of mobile agents. 
An agent is a small piece of software embedded in network element, functionally independent and can act upon their environment. Situatedness-Based Knowledge Plane is based on this principal, where network elements are agents which can act upon environment and making them as multi-agent system.

Agents can be classified based on situatedness of two kinds: First, \textit{Type of Situatedness}, this describes the type of neighborhood an network element is in. 
Second, \textit{Shape of Situatedness}, considers whether neighborhood is static or not, whether it includes all neighbors in the area or not. 
A perfect analogy is Open Shortest Path First (OSPF) protocol \cite{moy1997ospf}, but without wasting resources by scatter highly dynamic information over entire network. 
Each agent builds a primeval situated view of its environment by collecting control data from its hardware layer by setting sensors on each interface to sense variations of parameters. 
This control data is built by exchanging intermittent knowledge messages with its nearest neighbors. 
These knowledge messages are represented as facets. A facet represents a knowledge associated with a point of view. For instance, collected state of links is a facet. 

Situatedness-Based AI plane is based on event loop. 
A simple loop of network monitoring or logging protocol reads information from control algorithms (information such as load of interfaces, attacks etc) and an event is generated upon detecting a momentous change in the network. As depicted in Fig. \ref{fig:SituatednessBasedKP}, each network element (router in this case) is exchanging knowledge messages to built a extended view of the network. For variety of applications, this extended view can be accompanied with global information such as static topology conjunction with routing metrics. 

Although authors demonstrated promising results but there are several loopholes in that paradigm. First, there is no mention about the security of data being exchanged in knowledge messages. This particularly becomes important in federated networks \cite{FederatedNetworks}, where trust is a major issue for network elements in exchanging information. 
Second, this approach assumes that all network elements has same number of interfaces in order to calculate Computational Load Overhead (CLO) \cite{bullot2008situatedness}. 
This specifically becomes challenging to calculate CLO for a network element based on the number of information units (An information unit represents one item of control data by an interface of network element). Third is the assumption of network topology being infinite and acyclic: the arises a problem where a network element decision making process either has to work extra to filter out the redundant information or to process them, ultimately increasing latency, memory and CPU usage.

\subsection{NetQuery \cite{shieh2011netquery}} The biggest challenge in information gathering for AI plane is the security. 
Often times, information is required about the network elements, where administrative domain may have restrictive policies on disclosing network information. 
This prevents network elements to disseminate the information efficiently. 
To address this problem NetQuery \cite{shieh2011netquery}  applies \textit{trustworthy computing techniques} to ease reasoning in terms of trustworthiness of information belongs to AI plane. 
%This is the first AI plane focused on federated networks \cite{FederatedNetworks}. 
%Networks in federated environment might be reluctant to share the gathered information, 
NetQuery respects the disclosure of information such as routing tables, neighbor list etc while process and analyzes it.
%In AM module, NetQuery process and analyzes information such as routing tables, neighbor list etc and simultaneously respect the information disclosure of the participants, which might be reluctant to share it in federated environment.  
Prior to NetQuery, no AI plane was considering the possibility of federated networks security problem. 
Internet in one big picture is a collection of different ISPs (Internet Service Provider) networks interconnected on different protocols. 
An ISP prices a connection based on the path and performance they advertise. 
Cost of routing traffic differs on different paths and incenting operators \cite{goldberg2008rationality}. 
This motivated to evolve NetQuery, offering ISP to advertise their quality of network autonomically while maintaining the integrity of the information.

NetQuery starts by checking adjacent nodes network element entries (routing and forwarding table entries) and making sure that backup paths are only used at appropriate times. 
After analyzing the network element entries, NetQuery AI module makes a decision to forward traffic to minimal AS (Autonomous System) length. 
This AS length is BGP (Border Gateway Protocol) \cite{rekhter2005border} reported data and verified by traceroute \cite{malkin1993traceroute}. 
There is a good chance that there would be multiple paths with same AS length, network elements advertise this information via NetQuery using a reasoning process that observe the network topology. 

As shown in Fig. \ref{fig:NetQuery}, AI plane makes the network topology information available to applications in order to determine if the network is displaying coveted properties. 
Although, physically AI plane is federated, logically a global AI plane assimilates all properties across several managerial domains. Applications use AI plane to get information about any new or old participating network. Since, networks typically prohibits other networks to gain direct access, an abstraction layer \textit{sanitizers} are used to execute queries authorized by network admin to get those data sets. 
These queries are distinguished by a unique TID (tuple ID) and stores properties as pairs and associated metadata called factoid \cite{shieh2011netquery}. 
In order to provide interoperabilities between network elements, NetQuery uses schemas. Each schema represents the set of properties that a given type of network element must provide \cite{shieh2011netquery}.

%NetQuery uses a principal to symbolize a producer and consumer of query by sanitizers, this principal is represented by a unique public/private key pair generated independently. These principals are used for making control decisions. NetQuery records two control decision metadata for every factoid it stores. First, principal which generates the factoid. Second, which defines the factoid. 

Although NetQuery bring intelligence in network elements via AI plane, there are two major drawbacks which are left undiscussed. 
First, ISPs pursue to reduce the AS length but the paucity of establishing many direct interconnections prohibits it. 
This forces provider to either buy service from other providers to have AS length or embolden them to engage in peering. 
Second, each NetQuery independently defines set of rules it trusts. 
Since AI plane can collect information from various sources, burden of filtering such information comes on applications.  

\begin{figure}[t]
    \centering
    \includegraphics[width=0.75\linewidth]{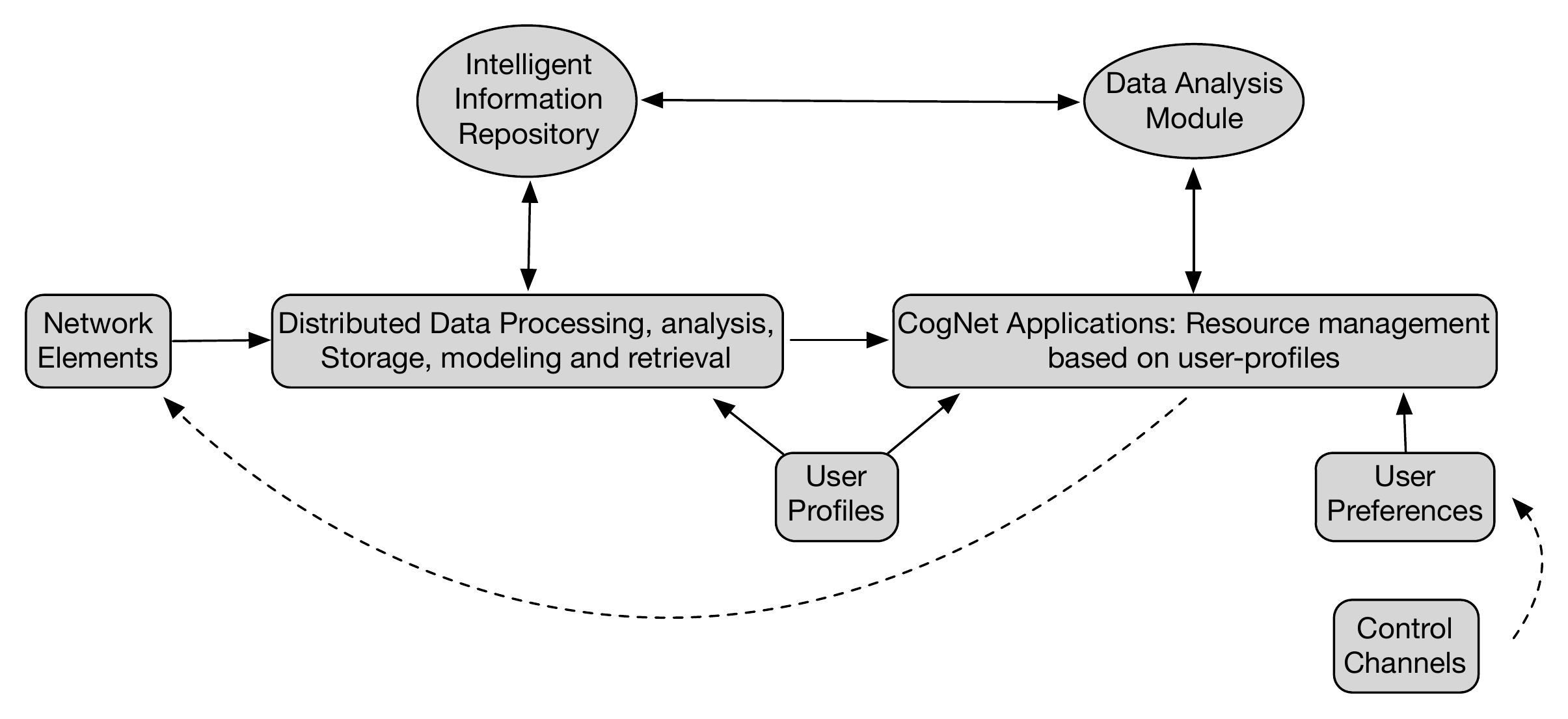}
    \captionsetup{font=scriptsize}
    \caption{\textbf{CogNet Architecture.}}
    \label{fig:CogNet}
\end{figure}

\subsection{A Cognitive Complete Knowledge Network System \cite{manoj2008cognet} (CogNet)} Over a period of time networks have evolved to provide ample amount of services to users. 
As \textit{Clark at el.} mentioned and true till date, it is becoming hard to manage and control these services. 
CogNet motivation is to address this problem. 
CogNet is service based and aimed to optimize one system. CogNet aids to identify the disparity in the awareness of network attributes such as: interference conditions, usage patterns etc. 
CogNet learns to optimize system through observing those disparities via extracting useful information from the network attributes. Cross-layer nature of CogNet bridges all layers (OSI Layers) of the protocol stack and notably addresses higher layer optimization.   

\begin{figure}[t]
    \centering
    \includegraphics[width=0.6\linewidth]{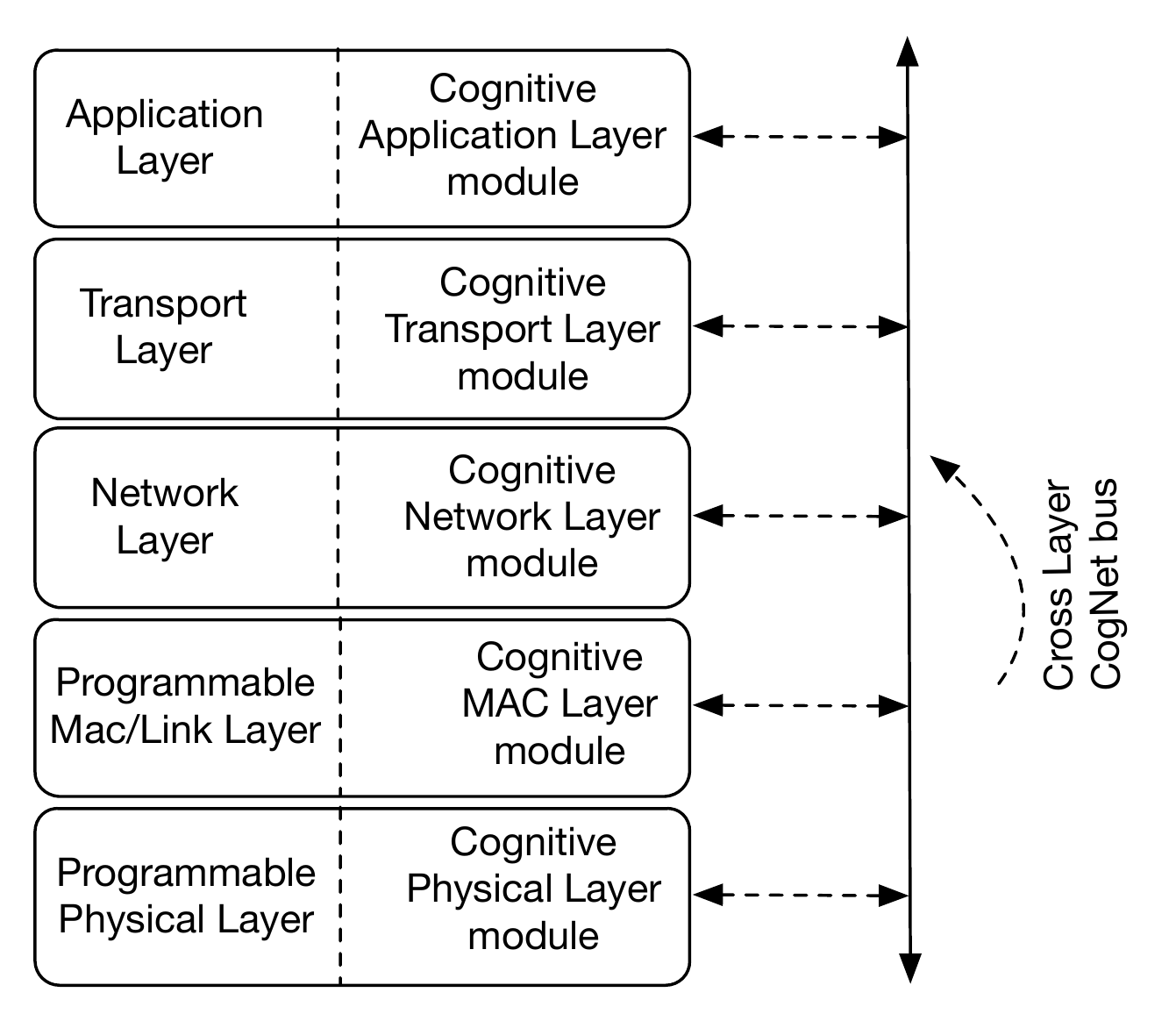}
    \captionsetup{font=scriptsize}
    \caption{\textbf{CogNet bus.}}
    \label{fig:CrossLayerCognetBus}
\end{figure}

CogNet defines cognitive agents sitting on every layer of protocol stack and responsible for gathering layer specific information (Higher layer:jitter, throughput etc. Lower layer:noise level, signal to noise ratio etc) from their respective layer of operation, and control the behavior of the attributes (attributes are entities of operation, for example: TCP in transport layer, a sensor node or a wireless radio). 
These attributes are key elements of CogNet and \textit{spatiotemporarily} stored in an intelligent repository. 
These attributes are transferred to AI plane via cross-layer \textit{CogNet bus}, where they are stored in a local repository. 
Main component of AI plane is a function called CogExec. CogExec applies learning algorithms to cull behavioral models of attributes. Learning algorithms in CogExec are of either short term where learning happens in individual layer or long term learning of overall system. Short term learning information is exchanged among layers via CogNet bus while long term learning information is stored in distributed intelligent repository. 
As soon as an application request arrives, AI plane executes learning algorithms for joint optimization and resource allocation. This helps to select the appropriate parameter and reconfigure protocols at each layer. 

Fig. \ref{fig:CogNet} shows fully distributed CogNet architecture. 
Every layer require to have a cognitive module to control, act and gather information and control the protocol parameters within that layer. Reason behind this layered structure of cognitive modules ( Fig. \ref{fig:CrossLayerCognetBus}) is for efficiently handling each layer's protocol parameter behavior, semantic interpretation of network events and actions taken. 
In addition, cognitive modules at each layer makes the joint optimization, static and dynamic resource allocation possible with the help of past history of user, device and network information. Each layer's cognitive module communicate via CogNet bus to exchange information. 
As Fig. \ref{fig:CogNet} shows, CogNet bus places a mechanism to exchange this cognitive information to achieve aforementioned cross layer tasks. CogNet bus must be lightweigth and format for information must be regonized on all layers. 
After cognitive information is received by each layer, AI plane (aka Cognitive plane) translates this information into end goals and responsibilities required for each layer. 
Each layer's cognitive module report their observation in a local repository and CogExec builds an interactive model to extract useful information to determinal final objective.

Main takeaway from CogNet is that it works well in heterogeneous networks by independently optimizing each layer and as a result empowering the entire system. It actively amasses, processes and hypothesizes information from a wide variety of sources for maintaining and dispersing context awareness in which users interact. However, authors excluded some fundamental flaws. First is storing information in intelligent information repository in long term learning, authors failed to highlight the issue of changing attributes during joint optimization. How changed attribute will affect CogExec and time elapsed between storing the new attribute in intelligent information repository? How protocol reconfiguration on a given layer will be affected? Is there any caching mechanism possible to overcome this issue?

\section{Network Level AI Plane}
Although inter-working of networks seems easier but mostly enables best effort data transport. In most cases any network consists of small heterogeneous networks interconnected via different protocols, however their control and management planes are often not compatible. This arises a need for an intelligent AI plane which can optimize entire network and not just one entity of it. This section explains such AI planes.

\subsection{Sophia \cite{wawrzoniak2004sophia} - 2003} This AI plane was influenced by an internet query processor PIER \cite{harren2002complex} and IrisNet \cite{nath2003irisnet}. 
Motivation behind Sophia was to manage several distributed machines worldwide such as PlanetLab \cite{peterson2003blueprint}. 
Prior to Sophia, in order to perform operations such as manage, store, sort, discover and aggregate information was challenging, specifically in terms of poor scaling and impoverished query language. 

Sophia is a distributed Prolog system \cite{clocksin2012programming} for information about networks. 
Sophia operates in three main functions. 
First function relates to set of sensors distributed in networks, these sensors report data about a particular node (memory usage or load on that particular node) or entire network view (for example, reachability to other nodes).
Sophia adapted a decentralized management system providing upper hand in managing and controlling a complex networks. 
Sensors are used to collect information about other sensors, feed those information to a distributed expression evaluator (called functor) and make conclusions about that information. 
These conclusions were used by actuators to take actions towards the final objective. 
Sophia decoupled high-level instructions from control and management plane and brought into their AI plane for control decisions.
%Main contribution of Sophia was to separate high-level instructions for networks from low-level instructions. For example: Prior to Sophia, in order to gather information about jitter based on location, network admin used to collect data with additional policies.  the \texttt{eval} functor would run 
%\texttt{eval(jitter env(node(<node-id>), <location>))}
In addition, Sophia concentrated on optimizing the AI plane by optimizing query computational latency\cite{wawrzoniak2004sophia}.

Although Sophia was succesful to implementing an intelligence in internet query, it lacks to define procedures how to use it efficiently for better management. 

\subsection{4D Approach \cite{greenberg2005clean} - 2005} Main motivation behind this approach was to ease the processes of \textit{programming} the network. Specifically it is based on three design principles: a) Network-level objects b) Network-wide views c) Direct Control. 
First two are for AM module (gather data and analyze them), while the third one belongs to AI module. 
Network-level objects specifies low-level configuration commands on network elements.  
Network-wide views refers to the snapshot of the state of each network element \footnote{Network-wide views reflects the present state of the data plane, including information about device such as name, physical attributes, resource limitations etc}. 
In other words, Network-Wide views should mirror the current state of the data plane and information about each network element. 

\begin{figure}[t]
    \centering
    \includegraphics[width=0.7\linewidth]{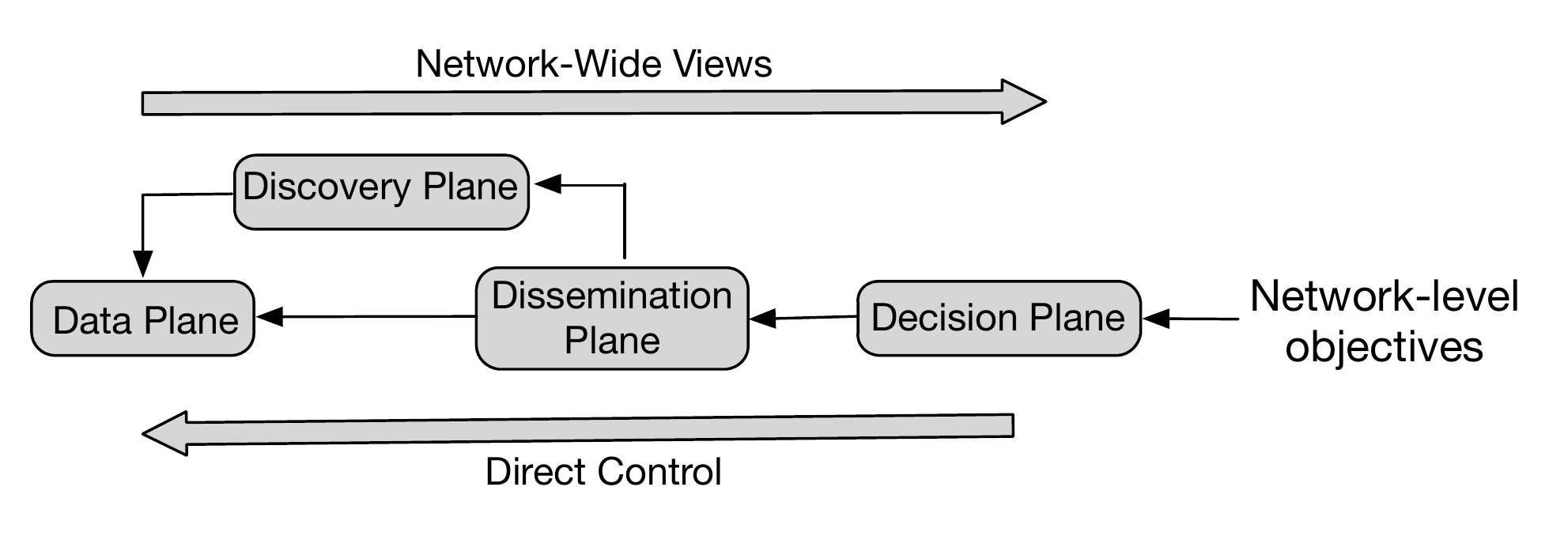}
    \captionsetup{font=scriptsize}
    \caption{\textbf{Network-level objectives.}}
    \label{fig:4D}
\end{figure}

There are three main components of 4D approach depicted in Fig. \ref{fig:4D}. First, Decision Plane, it dissociates load balancing, reachability, interface configuration, security and network control from management plane and operates in real-time on the network-wide view of the topology. Decision plane acts like an AI module which converts the network-level objectives (reachability, load balancing goals etc) directly into rules to be configured in data plane (forwarding entries, packet filters, queues etc). 
Second is dissemination plane, acts like a communication bus from data to decision plane and vice versa. 
Although the control plackets travels through the same path as data packets but they are kept separately so that they can be available even in absence of any configuration, unlike in today's networks where control plane packets needs to be prioritized to establish communication beforehand. 
Third is discovery plane, responsible for locate physical components in the network and create and symbolize them by logical identifiers. This logical identifiers assists discovery plane to create a network-wide view \cite{greenberg2005clean}. 
A perfect example would be neighbor discovery in network to find out: type of interfaces and their state on the network element, how many forwarding entries can be held by a network element, type of device connected to network element etc. 
Decision plane uses this information and creates a network-wide objectives, unlike in today's IP networks where neighbor discovery is done by configuration commands.

Aforementioned three principles are materialized by 4D architecture. Decision plane, with the help of discovery plane, operates on network topology (network-wide view) and traffic to get network-level objectives. Although, 4D approach showed several advantages mentioned in \cite{greenberg2005clean}, it fails to address some key issues. First, the communication channel from decision to data plane is being shared by data packets too. Although authors have mentioned about keeping them separate in a common communication path, however it is hard to project the flow of control packets as network starts to grow. Second important point is addressing the problem of limited forwarding entries. In case of limited forwarding entries, which traffic will take precedence over others? How do we prioritize them?

\subsection{\textit{iPlane} \cite{madhyastha2006iplane}} 
iPlane is scalable service predicting path performance in overlay networks. 
Many overlay networks are hazy to their applications. 
For example, overlay networks in CDN (Content Delivery Networks \cite{CDNnetwork}) like Coral \cite{freedman2004democratizing}, CoDeeN \cite{wang2004reliability} and Akamai maintains replica of connections by each client and redirect them based on best performance. 
iPlane was introduced to reduce the gap between overlay networks and their applications by providing an AI plane as a service. Although, there are several existing prediction services such as: RON \cite{andersen2001resilient} and \texttt{S$^3$} \cite{lee2005measuring}, but they don't solely focus on intra-overlay paths. iPlane makes prediction for arbitrary internet path based on the accurate estimation of performance metrics such as latency, loss rate etc. 
iplane unitedly co-ordinates among network elements and generates and maintains a detailed and comprehensive outline of a network. While co-ordinating iplane does two kinds of measurements: Active Measurement: determines the attributes of network elements and the links joining them. 
Opportunistic Measurement: monitoring the actual data transforms moving end to end. 

First step is, mapping the network. 
iPlane's elementary tool for finding network topology is \texttt{traceroute} \cite{malkin1993traceroute}, which determines the forward path from the probing network element to the destination. 
iPlane takes snapshots of all the routing prefixes learned by network elements and carefully goes through one by one prefix and under each prefix starts probing .1 address, which is expected to respond either ICMP \cite{deering1991icmp} or UDP (User Datagram Protocol \cite{postel1980user}) probe (For UDP probing iPlane uses Mercator \cite{govindan2000heuristics} technique). 
It is assumed that .1 address is either a router and most likely to respond. 
In order to compress the prefixes, iPlane uses BGP atom \cite{broido2001analysis}, which generates a crammed list of prefixes. 
This list is called \textit{probe target list}. 
A list of interfaces will be populated by traceroute from source to destination. 
It should be noted that interfaces on the same router would have akin behavior as they belong to the same network element. 
In order to avoid that repetition and make topology more condensed, iPlane segregate interfaces in clusters, we can think clusters as an autonomous system. 
To arrange those clusters geographically, iPlane concludes the DNS \cite{mockapetris2004rfc} names entrusted to interfaces by using \texttt{Rocketfuel} \cite{spring2002measuring} and \texttt{Sarangworld Project} \cite{SarangWorld} \footnote{For several reasons DNS resolution might fail, in this case iPlane uses its automated algorithm based on ICMP echo, please refer \cite{madhyastha2006iplane}}. 
After gathering data about routing topologies of interfaces belong to network elements. To analyze the data, iPlane developed a frontier algorithm to assign tasks to vantage points. Frontier algorithms only measure points at the border of the link. Frontier algorithms perform measurements along traversing the link simultaneously. 

\begin{figure}[t]
    \centering
    \includegraphics[width=0.75\linewidth]{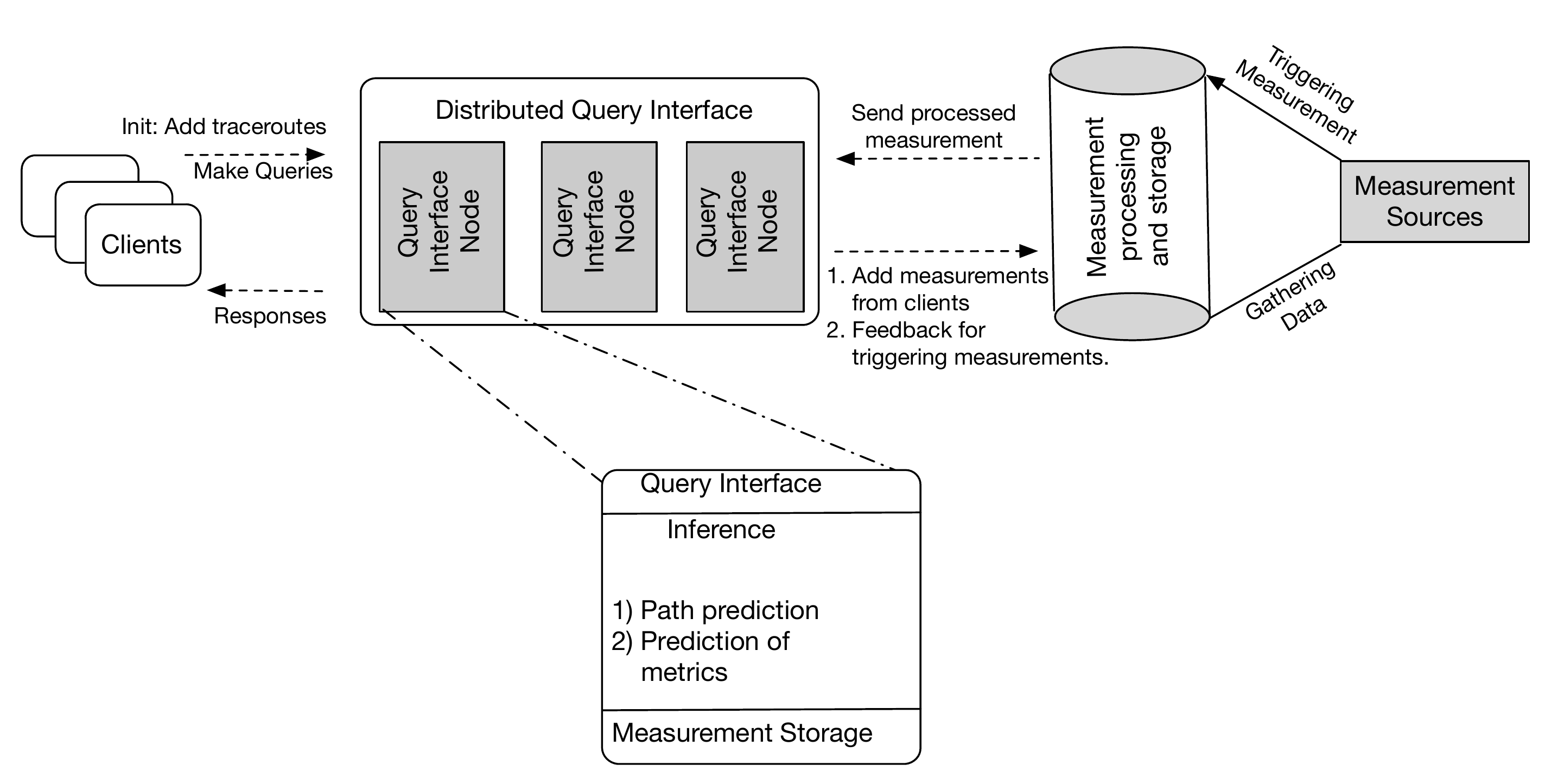}
    \captionsetup{font=scriptsize}
    \caption{\textbf{Distributed Query Interface and response between two end-hosts.}}
    \label{fig:iPlane}
\end{figure}

In order to make intelligent control decisions based on gathered data by forming clusters and measurements, iPlane run \textit{Performance Prediction}, which is divided in two parts: 1) Predicting the forward and reverse path 2) Accumulates measured link-level properties to envision end to end path properties. For path prediction, iPlane uses a structural technique \cite{madhyastha2006structural} and by simply aggregating link-level properties, iPlane can estimate end to end properties. Various components and their interactions are shown in Fig. \ref{fig:iPlane}.

Although iPlane has vast number of applications, but it introduces a lot of issues. 
Authors \cite{madhyastha2006iplane} has mentioned one major issue is \textit{security}. 
While making clusters, iPlane allows untrustworthy network elements to participate and this can pollute the information set exchanged among clusters and within a cluster. 
There are other issues with iPlane which authors have not explained or included. 
First one is traceroute, authors in \cite{TracerouteIssue} have laid down some important limitations by traceroute, which could possibly be bottleneck here. 
Second issue is routing table, there are several network elements in overlay networks, which are layer 2 capable and not being able to have routing table will exclude them to have the cluster topology. 
For instance, VxLAN \cite{mahalingam2014virtual} combines two different network broadcast domains as one. 
So any packet initiated from one network will appear as the same subnet packet for network 2. 
This will make traceroute not including those VTEP (VXLAN Tunnel Endpoint) and hence this can pollute the topology and can hinder in formation of control decisions. 
Third major issues is the nature of routing protocol. For example, populating routing table based on one protocol is different than others. For instance, loop detection/avoidance in OSPF is different than BGP. 
This can seriously cause issues in populating prefixes and resulting to have a corrupt cluster. 

\subsection{COBANETS (Cognition-BAsed-NETworks) \cite{zorzi2016cobanets}} 

Variety of services are being deployed by providers to cope with the quality of service (QoS) given to customers. 
This is challenging due to lack of scalability in current heterogeneous networks.
For instance, authors in \cite{boccardi2014five, chin2014emerging, demestichas20135g} defined \textit{heterogeneity} in traffic and \textit{scalability} in number of network functions and parameters in a network element, can scarce resources such as Quality of Experience (QoE) , bandwidth and energy. 
This arises an absolute need for an effective resource management. 
COBANETS was built to address those issues. 
Final objective of system wide optimization in COBANET is achieved by introducing a Generative Deep Neural Networks (GDNN). 
GDNN makes predictions and detects anomaly by minimizing the error between inner representation of the input and observed data given by the hidden layer. 

Architectural wise, GDNN has three main components. First is \texttt{abstraction and optimization}: Essential components of COBANETS are generative models providing a descriptive portrayal of crucial elements and functionalities of a network, for instance: traffic originators, physical and access medium protocols. 
There are more than one instance of GDNN, capturing the linkage between other parameters belong to the same protocol layer. 
These generative models are used to forecast the recent offered traffic in future or to train classifier to extract detailed context information. 
An example of context information would be data flows generated by types of applications, operation environment like outdoor, urban, vehicular etc. 
This context information is used in optimizing functionality of network elements such as: caching, handover and transit rate etc. 
Second is \texttt{integration of generative models}: this assimilation is heavily based on \cite{srivastava2012multimodal}. 
A possible solution of integrating generative models would be concatenate and train them together. 
This introduces an additional task of recreating those composite input and learning useful information among those generative models. 
Another approach is to categorize those generative models based on layer-specific indices and train them together. 
Third is \texttt{optimization among flows}: this step is responsible to optimize multiple functionalities within a single network element. Main difference between traditional cross layer optimization \cite{park2017cross} and COBANETS is that it doesn't assume any interdependencies among protocols. 
As an example, user watching videos on their smart phones in vehicular networks reflects a specific inter-relations among the categories of generated traffic by these devices, interference caused by other devices and access network characteristics. 
COBANETS produces generative models adept of these abundant correlations, hence optimizing strategies to that particular device or scenario. Finally \texttt{System Level Optimization}: In order to achieve system wide optimization, COBANET shall have a global scope rather than a single network element. 
Trained generative models are comprehensively optimized based on certain data such as: nature of the data, characteristics of the end user, link congestion etc. 
A pragmatic example of understand GDNN is in \cite{munaretto2015data}, in which GDNN is trained to understand a generative model of the size of encoded video frames by estimating the rate or distortion curve of every video sequence to design QoE aware resource allocation and admission control algorithms \cite{testolin2014machine}-\cite{ zanforlin2014ssim}. 

Despite GDNN architecture looks promising, but it has some caveats. Although \cite{zorzi2016cobanets} mentioned important ones but some are not covered by the authors. For instance, authors mentioned the optimization of data collection but transferring the data still posses some challenges about occupying network bandwidth and how their prioritization would be done. Other imperative one is the change in generative model, authors failed to mention about how the integration and training will adapt if generative models change during integration. 


\chapter{Future Challenges}
Despite the fact there numerous attempts to define and implement AI plane have been done. There are still unanswered important questions.

% \usepackage{multirow}
\begin{table}[]
\begin{center}
\caption{Brief summary of all AI planes}
\label{tab:AIPlanesSummary}
\begin{tabular}{|l|c|c|c|c|c|}
\hline
\rowcolor[HTML]{C0C0C0} 
\multicolumn{1}{|c|}{\cellcolor[HTML]{C0C0C0}Type}                                                      & Name                                                                         & Network Type                                                                & Methods                                                                                 & \begin{tabular}[c]{@{}c@{}}Layers of\\ Operations\end{tabular}                                                & Year          \\ \hline \hline
\multicolumn{1}{|c|}{}                                                                                  & \begin{tabular}[c]{@{}c@{}}Situatedness-based\\ Knowledge Plane\end{tabular} & Large Scale                                                                 & \begin{tabular}[c]{@{}c@{}}Administered\\ Un-administered\end{tabular}                  & \begin{tabular}[c]{@{}c@{}}Data Link Layer\\ Network Layer\\ Transport Layer\\ Application Layer\end{tabular} & 2008          \\ \cline{2-6} 
\multicolumn{1}{|c|}{}                                                                                  & NetQuery                                                                     & Large Scale                                                                 & \begin{tabular}[c]{@{}c@{}}Administered\\ Un-administered\end{tabular}                  & Network Layer                                                                                                 & 2011          \\ \cline{2-6} 
\multicolumn{1}{|c|}{\multirow{-3}{*}{\begin{tabular}[c]{@{}c@{}}System Level\\ AI Plane\end{tabular}}} & CogNet                                                                       & \begin{tabular}[c]{@{}c@{}}Wireless\\ (Cognitive Radio)\end{tabular}        & \begin{tabular}[c]{@{}c@{}}Short and long\\ term learning (Prime)\end{tabular}          & \begin{tabular}[c]{@{}c@{}}Data Link Layer\\ Transport Layer\end{tabular}                                     & 2015          \\ \hline
                                                                                                        & \textit{Sophia}                                                              & \textit{Large Scale}                                                        & \textit{Administered}                                                                   & \textit{\begin{tabular}[c]{@{}c@{}}Data Link Layer\\ Network Layer\end{tabular}}                              & \textit{2003} \\ \cline{2-6} 
                                                                                                        & 4D Approach                                                                  & \textit{Large Scale}                                                        & \textit{\begin{tabular}[c]{@{}c@{}}Administered\\ Un-administered\\ Prime\end{tabular}} & \textit{Network Layer}                                                                                        & \textit{2005} \\ \cline{2-6} 
                                                                                                        & iPlane                                                                       & \textit{\begin{tabular}[c]{@{}c@{}}Peer-to-peer\\ Large Scale\end{tabular}} & \textit{\begin{tabular}[c]{@{}c@{}}Administered\\ Un-administered\\ Prime\end{tabular}} & \textit{\begin{tabular}[c]{@{}c@{}}Network Layer\\ Transport Layer\\ Application Layer\end{tabular}}          & \textit{2006} \\ \cline{2-6} 
\multirow{-4}{*}{\begin{tabular}[c]{@{}l@{}}Network Level\\ AI Plane\end{tabular}}                      & COBANETS                                                                     & \textit{\begin{tabular}[c]{@{}c@{}}CDN\\ 5G\\ Large Scale\end{tabular}}     & \textit{\begin{tabular}[c]{@{}c@{}}Administered\\ Un-administered\\ Prime\end{tabular}} & \textit{\begin{tabular}[c]{@{}c@{}}Network Layer\\ Transport Layer\\ Application Layer\end{tabular}}          & \textit{2016} \\ \hline
\end{tabular}
\end{center}
\end{table}

\begin{table}[]
\centering
\caption{System Level AI Planes}
\label{tab:AIPlanesComparison1}
\begin{tabular}{c|c|c|c|}
\cline{2-4}
\multicolumn{1}{l|}{}                                                                                                       & \cellcolor[HTML]{C0C0C0}\textbf{\begin{tabular}[c]{@{}c@{}}Situatedness-Based\\ Knowledge Plane\end{tabular}} & \cellcolor[HTML]{C0C0C0}\textbf{NetQuery}                       & \cellcolor[HTML]{C0C0C0}\textbf{CogNet}                                 \\ \hline
\multicolumn{1}{|c|}{\cellcolor[HTML]{C0C0C0}\textbf{\begin{tabular}[c]{@{}c@{}}Application\\ Classification\end{tabular}}} & \begin{tabular}[c]{@{}c@{}}Open Loop\\ Closed Loop\end{tabular}                                               & \begin{tabular}[c]{@{}c@{}}Open Loop\\ Closed Loop\end{tabular} & \begin{tabular}[c]{@{}c@{}}Closed\\ Loop\end{tabular}                   \\ \hline
\multicolumn{1}{|c|}{\cellcolor[HTML]{C0C0C0}\textbf{Mapping Tool}}                                                         & \begin{tabular}[c]{@{}c@{}}Multi-Agent\\ System\end{tabular}                                                  & \begin{tabular}[c]{@{}c@{}}BGP \\ Traceroute\end{tabular}       & CogBus                                                                  \\ \hline
\multicolumn{1}{|c|}{\cellcolor[HTML]{C0C0C0}\textbf{\begin{tabular}[c]{@{}c@{}}Resolute\\ Language\end{tabular}}}          & Stack of Facet                                                                                                & Factoid                                                         & \begin{tabular}[c]{@{}c@{}}Parameters\\ of interest\end{tabular}        \\ \hline
\multicolumn{1}{|c|}{\cellcolor[HTML]{C0C0C0}\textbf{Motivation}}                                                           & \begin{tabular}[c]{@{}c@{}}Make Services\\ Scalable\end{tabular}                                              & \begin{tabular}[c]{@{}c@{}}Federated\\ Networks\end{tabular}    & \begin{tabular}[c]{@{}c@{}}Manage Control\\ Plane Services\end{tabular} \\ \hline
\multicolumn{1}{|c|}{\cellcolor[HTML]{C0C0C0}\textbf{Cross-Layer}}                                                          & Yes                                                                                                           & No                                                              & Yes                                                                     \\ \hline
\multicolumn{1}{|c|}{\cellcolor[HTML]{C0C0C0}\textbf{\begin{tabular}[c]{@{}c@{}}Data Gathering\\ Method\end{tabular}}}      & \begin{tabular}[c]{@{}c@{}}Knowledge\\ Messages\end{tabular}                                                  & Schema                                                          & \begin{tabular}[c]{@{}c@{}}CogNet\\ Repository\end{tabular}             \\ \hline
\end{tabular}
\end{table}

\section{Gathering and Transferring Data} The key enabler for AI plane is gathering and exchanging data. Finding the more relevant data, studying their granularity and other properties, provided the time elapsed in this discovery is minimal is an open issue for research. As more network elements are introduced, innovative methods would be needed to store, query and manage data.

\section{Synchronizing and Representing Data} Data gathered from different network elements can be of different format and can lead to polluting the control decision. Although few literature such as: \cite{whiteson2004adaptive} have solved it by having a pre-defined message (Load Update) format but this forces network elements to strictly follow that specific format. This arises a need for a standardized formatting scheme for AI plane. Author \cite{bouabene2010autonomic} addresses this issue by the help of an API (\texttt{getIDPinfo}) but this approach is very architecture specific. Lately authors in \cite{lopez2018machine} introduced an extra entity SDN (Software Defined Networking Plane) plane to solve this problem but this requires an extra hardware and communication between AI planes and SDN plane. To overcome all these limitations, a globally recognized format needs to be researched.

\section{Multi-level Objects Optimization Strategy}
An AI plane is made of several components. Although some components would work towards optimizing the entire network or system but some components might have selfish objectives. This issue can be resolved by innovative optimization strategies for Multi-level objects via defining some utility functions that accounts for optimizing multiple objectives combined. This probably will be coupled with prime learning to learn the best strategy for system or entire network.

\section{Security} Whatever the final goal of AI plane is, whether it is to manage the network, program the network or optimize it, a large amount of data is collected and machine learning algorithms are applied to it. This makes the confidentiality of data much more important then the traditional TCP/IP networks. For instance, by changing the behavior and inspecting how network reacts, it may be possible for hacker to get private information about other others \cite{acs2013cache}. An open challenge is to find the tradeoff point between secrecy and effectiveness in AI plane. Authors in \cite{conti2015can} have addressed this issue partially by considering \textit{de-anonymization} techniques and privacy attached via machine learning. 

\begin{table}[]
\centering
\caption{Network Level AI Planes}
\label{tab:AIPlanesComparison2}
\begin{tabular}{c|c|c|c|c|}
\cline{2-5}
\multicolumn{1}{l|}{}                                                                                              & \cellcolor[HTML]{C0C0C0}\textit{Sohpia}                                     & \cellcolor[HTML]{C0C0C0}\textit{4D Approach}                                                 & \cellcolor[HTML]{C0C0C0}\textit{iPlane}                                        & \cellcolor[HTML]{C0C0C0}\textit{COBANETS}                                \\ \hline
\multicolumn{1}{|c|}{\cellcolor[HTML]{C0C0C0}\begin{tabular}[c]{@{}c@{}}Application\\ Classification\end{tabular}} & \textit{\begin{tabular}[c]{@{}c@{}}Closed Loop\\ Open Loop\end{tabular}}    & \textit{\begin{tabular}[c]{@{}c@{}}Open Loop\\ Closed Loop\end{tabular}}                     & \textit{\begin{tabular}[c]{@{}c@{}}Open Loop\\ Closed Loop\end{tabular}}       & \textit{\begin{tabular}[c]{@{}c@{}}Closed Loop\\ Open Loop\end{tabular}} \\ \hline
\multicolumn{1}{|c|}{\cellcolor[HTML]{C0C0C0}Mapping Tool}                                                         & \textit{\begin{tabular}[c]{@{}c@{}}PIER\\ IrisNet\end{tabular}}             & \textit{BGP}                                                                                 & \textit{\begin{tabular}[c]{@{}c@{}}Traceroute\\ DNS\end{tabular}}              & \textit{GDNN}                                                            \\ \hline
\multicolumn{1}{|c|}{\cellcolor[HTML]{C0C0C0}\begin{tabular}[c]{@{}c@{}}Resolute\\ Language\end{tabular}}          & \textit{Prolog}                                                             & \textit{\begin{tabular}[c]{@{}c@{}}Network-wide\\ Views\end{tabular}}                        & \textit{Clusters}                                                              & \textit{\begin{tabular}[c]{@{}c@{}}Generative\\ Models\end{tabular}}     \\ \hline
\multicolumn{1}{|c|}{\cellcolor[HTML]{C0C0C0}Motivation}                                                           & \textit{\begin{tabular}[c]{@{}c@{}}Distributed\\ System Query\end{tabular}} & \textit{\begin{tabular}[c]{@{}c@{}}Ease the process\\ of Programming\\ Network\end{tabular}} & \textit{\begin{tabular}[c]{@{}c@{}}Optimizing\\ Overlay Networks\end{tabular}} & \textit{ImprovingQoS}                                                    \\ \hline
\multicolumn{1}{|c|}{\cellcolor[HTML]{C0C0C0}Cross-Layer}                                                          & \textit{No}                                                                 & \textit{No}                                                                                  & \textit{No}                                                                    & \textit{Yes}                                                             \\ \hline
\multicolumn{1}{|c|}{\cellcolor[HTML]{C0C0C0}\begin{tabular}[c]{@{}c@{}}Data Gathering\\ Method\end{tabular}}      & \textit{Caching}                                                            & \textit{\begin{tabular}[c]{@{}c@{}}Dissemination\\ Plane\end{tabular}}                       & \textit{\begin{tabular}[c]{@{}c@{}}Distributed Query\\ Interface\end{tabular}} & \textit{\begin{tabular}[c]{@{}c@{}}Measurement\\ Sources\end{tabular}}   \\ \hline
\end{tabular}
\end{table}

\section{Adaptation of new Machine Learing Algorithms} Network evolved as a hardware centric engineering. Researchers always focused on building hardware to address the issue of increasing users and applications. With the rise of SDN paradigm \cite{dabbagh2015software}, software components in networking have become an important part. Introducing AI plane paradigm aggravated it and required new set of skills, particularly Machine Learning (ML) algorithms. As new ML algorithms are developed, AI plane needs to shift its model to adopt those. For example, graph and trie algorithms \cite{nilsson1998implementing} are used to represent network topologies, a crucial part of network performance and applications. New ML algorithm obsolete those algorithms by proposing new and efficient algorithm to determine network topology. AI planes need to embrace those ML algorithm to calculate network topology efficiently.


\chapter{Conclusion}
With the increasing growth of networks and their need to manage, program and optimize, AI plane is going to be the future of network infrastructure. Several latest paradigms such as NFV, 5G are already exploring machine learning algorithms \cite{peng2015system} \cite{jiang2017machine}. It is certain that AI plane will play an important part in it. 

Table \ref{tab:AIPlanesSummary} represents the AI planes comparison based on the network type, methods of ML algorithms classification, layers of operation and whether AI plane has inbuilt security or not. In order to further classify, Table \ref{tab:AIPlanesComparison1} and \ref{tab:AIPlanesComparison2} presents comparison based on application set AI plane fits in, mapping tool to construct network view, resolute language which serves as a common interface between human and AI module, motivation behind each AI plane, cross layer functionality and their data gathering methods.

Comparison of all AI planes is in the order of time they were proposed. We identified flaws which were left by authors and compared the AI planes irrespective of their networks types. We also proposed the future scope of research in AI planes. Although AI planes have evolved significantly, there are still open challenges need to be addressed. 

% \usepackage{booktabs}

\pagebreak

\nocite{*} % to test all bib entrys
\bibliographystyle{unsrt}
\bibliography{main} % file mwe.bib
% Adding a bibliography if citations are used in the report
%\bibliographystyle{plain}
%\bibliography{main.bib}
% Adds reference to the Bibliography in the ToC
\addcontentsline{toc}{chapter}{\bibname}

\end{document}